\documentclass[runningheads]{llncs}
\usepackage{graphicx}
\usepackage{xcolor}
\usepackage{hyperref}
\begin{document}
\title{Intervertebral Disc Labeling With Learning Shape Information, A Look Once Approach}

\author{Reza Azad\inst{1} \and
Moein Heidari\inst{2} \and Julien Cohen-Adad\inst{3, 4, 5} \and Ehsan Adeli\inst{6} \and Dorit Merhof\inst{1,7}}
%First names are abbreviated in the running head.
\institute{Institute of Imaging and Computer Vision,
RWTH Aachen University, Germany\and
School of Electrical Engineering, Iran University of Science and Technology, Iran, moein\_heidari@elec.iust.ac.ir \and 
Functional Neuroimaging Unit, CRIUGM, University of Montreal, Canada \and
NeuroPoly Lab, Institute of Biomedical Engineering, Polytechnique Montreal, Canada \and
Mila, Quebec AI Institute, Canada, \email{jcohen@polymtl.ca} \and Stanford University, \email{eadeli@stanford.edu} \and
Fraunhofer Institute for Digital Medicine MEVIS, Bremen, Germany\\\email{\{azad,dorit.merhof\}@lfb.rwth-aachen.de}}

\maketitle              % typeset the header of the contribution

\begin{abstract}
Accurate and automatic segmentation of intervertebral discs from medical images is a critical task for the assessment of spine-related diseases such as osteoporosis, vertebral fractures, and intervertebral disc herniation. To date, various approaches have been developed in the literature which routinely rely on detecting the discs as the primary step. A disadvantage of many cohort studies is that the localization algorithm also yields to false positive detections. In this study, we aim to alleviate this problem by proposing a novel U-Net-based structure to predict a set of candidates for intervertebral disc locations. In our design, we integrate the image shape information (image gradients) to encourage the model to learn rich and generic geometrical information. This additional signal guides the model to selectively emphasize the contextual representation and to supress the less discriminative features. On the post-processing side, to further decrease the false positive rate, we propose a permutation invariant 'look once' model, which accelerates the candidate recovery procedure. In comparison with previous studies, our proposed approach does not need to perform the selection in an iterative fashion. The proposed method was evaluated on the spine generic public multi-center dataset and demonstrated superior performance compared to previous work. We have provided the implementation code in \href{https://github.com/rezazad68/intervertebral-lookonce}{\textcolor{red} {github}.}

\keywords{Deep learning  \and intervertebral disc labeling \and look once \and shape feature.}
\end{abstract}

\section{Introduction}
The human vertebral column consists of 33 individual vertebrae stacked on top of each other and connected through the ligaments and intervertebral discs (IVDs). The vertebral column is divided into cervical, thoracic, lumbar, sacral and caudal vertebrae \cite{azad2021stacked}. Each of these regions performs a vital function in the human body including, absorbing shock, load breathing, protection of the spinal cord, controlling load through the vertebral column, and so on \cite{al2022transfer}. More precisely, the IVDs act as cushions of fibrocartilage and as principal joints between vertebrae and they absorb the stress and shock the body sustains during motion and allow the spine to be flexible while preventing the vertebrae from grinding against one another. Disruption in any of the vertebral discs through aging, degeneration, or injury will result in an alteration in the corresponding disc's properties along with flaws in mechanical functionalities of adjacent tissues \cite{urban2003degeneration}. As a consequence, location and segmentation of intervertebral discs is a crucial task for spine disease diagnosis and provides versatile information in the quality of treatment procedure. To this end, various semi-automated and automated techniques have been proposed in the literature.
Like other medical image processing researches \cite{azad2019bi,bozorgpour2021multi,feyjie2020semi,asadi2020multi,azad2022medical,reza2022contextual,azad2021deep,azad2021texture,azad2020attention} , these methods can be divided into two taxonomies: hand-crafted methods and deep learning-based approaches. As an example for hand-crafted dissertations, Cheng et al. \cite{chen2015localization} proposed a two-step approach where they first localize the center of each IVD by adapting a data-driven estimation framework \cite{chen2014automatic}
and, then, segment IVDs by classifying image pixels around each disc center as either foreground (disc) or background. 
Glocker et al. \cite{glocker2012automatic} utilized a regression forest and a probabilistic graphical model to detect
and localize intervertebral discs from CT scan images. A polynomial iterative randomized Hough transform approach to segment the spine and intervertebral discs was proposed in \cite{badarneh2021semi}. Irrespective of the good performance of these traditional methods, in some cases they intrinsically render poor performance when compared to deep learning-based methods \cite{ayed2011graph,chen2015localization}.
Recent advances in deep learning have facilitated investigation of robust intervertebral
disc labeling \cite{cheng2021automatic,vania2021intervertebral,chen2021study}.
In \cite{ji2016automated} the authors proposed to use a standard CNN for IVD segmentation. Dolz et al. \cite{dolz2018ivd} proposed an architecture called 'IVD-Net' to leverage information from multiple image modalities for inter-vertebral disc segmentation by adopting a U-Net-like architecture. In a recent article Vania et. al. \cite{vania2021intervertebral} developed a method which builds upon mask-RCNN and formulated a multi-optimization training system at a different stage to increase the computational efficiency. In another approach \cite{wimmer2018fully}, a cross-modality method for detecting both vertebral and intervertebral discs on volumetric data has been proposed. This approach utilizes a local entropy-based texture model to localize the sacral region. Then, using three-disc entropy models, detected positions are aligned and further refined by taking into account the intensity match between regions and a spinal column template. A transfer learning-based approach is utilized by \cite{mbarki2020lumbar}. In this work, a 2D convolutional structure is exploited to detect the lumbar disc from axial images. Their proposed network uses the strength of the U-Net structure with a VGG backbone to produce a spine segmentation mask. Then, the segmented regions are used to calculate the herniation in lumbar discs.
The authors of \cite{rouhier2020spine} combine a fully convolutional network  with inception modules to localize and label intervertebral discs. Azad et al. \cite{azad2021stacked} reformulated the semantic vertebral disc labeling using pose estimation and utilized an hourglass neural network to semantically label the intervertebral discs. \\
The main limitation of the reviewed methods is their dependency on the regular CNN learning strategy (learning texture, shape, colour) which is not optimal for labelling anatomical structures such as intervertebral discs and usually produces both false positive (FP) and false negative (FN) detections. To overcome this issue, we propose to incorporate shape information within the learning process. This additional signal guides the model to selectively emphasize the contextual representation, magnifies the structural regions and supresses the less discriminative features (e.g. color, texture). 

Moreover, a principal limitation of many cohort studies is that, as they utilize the local maximum technique to locate the position of the vertebral discs in 2D space on top of the prediction masks, they encounter a substantial false positive rate. Exhaustive search tree \cite{azad2021stacked}, template matching \cite{ullmann2014automatic} and point coordinate condition \cite{rouhier2020spine} are among the popular algorithms proposed to eliminate the FP rate. However, these approaches usually lack computational efficiency and render a poor candidate recovery. Therefore, a general method is required to handle this challenge. In this work, we propose to mitigate this limitation by bolstering the post-processing step in the intervertebral disc labeling procedure. The main idea is that, inspired by the idea of YOLO \cite{redmon2016you}, we propose a permutation invariant 'look once' model to increase the True Positive (TP) rate while reducing the FN detection. We re-formulate the problem by a modified version of the PointNet model \cite{qi2017pointnet} which is invariant to certain geometric transformations (e.g. rotation). To the best of our knowledge, this is the first post-processing algorithm that processes the whole prediction in one step without any iteration ('look once'). Our contributions are as follows :

$\bullet$ Adapting U-Net structure for semantic intervertebral disc labeling

$\bullet$ Incorporation of shape information to further boost model performance

$\bullet$ A permutation-invariant post-processing approach to reduce the FP rate

$\bullet$ Publicly available implementation source code (once accepted)

\section{Proposed Method}
Our proposed method consists of two stages. In the first stage we utilize a U-Net-based structure to detect and predict semantic labeling for each intervertebral disc location. In the second stage, we propose a deep permutation invariant 'look once' model to refine the prediction results and eliminate the FP candidates. In the next subsections, we will discuss each phase in more detail.

\subsection{Semantic Intervertebral Disc Labeling}
The concept of the proposed method is depicted in Fig. 1.
In our novel design, we incorporate the shape information (gradient of the input image) as an additional signal to encourage the model to learn contextual and geometric information. To this end, we form a pyramid representation using the multi-level description resulting from each block of the encoder (U-Net encoder $E$ parametrized with $\theta$) module: $P = \{f_j = E(x, \theta), j=0,1, ... L\}$, where $L$ is the number of pyramid levels. Next, we propose a shape attention module. Our attention module (Fig. 2) uses the global representation of each feature map alongside the shape description to selectively emphasize the contextual representation and supress the less discriminative features. 
To this end, for each level of the pyramid, we learn the channel-wise recalibration parameters ($w_j^f$) and spatial recalibration parameters ($w_{sp}$) from the shape feature description ($sf$):

\begin{figure}[t]
\label{fig:fig1}
%\vspace{-5mm}
	\centering
	\begin{tabular}{cc}
		\includegraphics[width=1\textwidth]{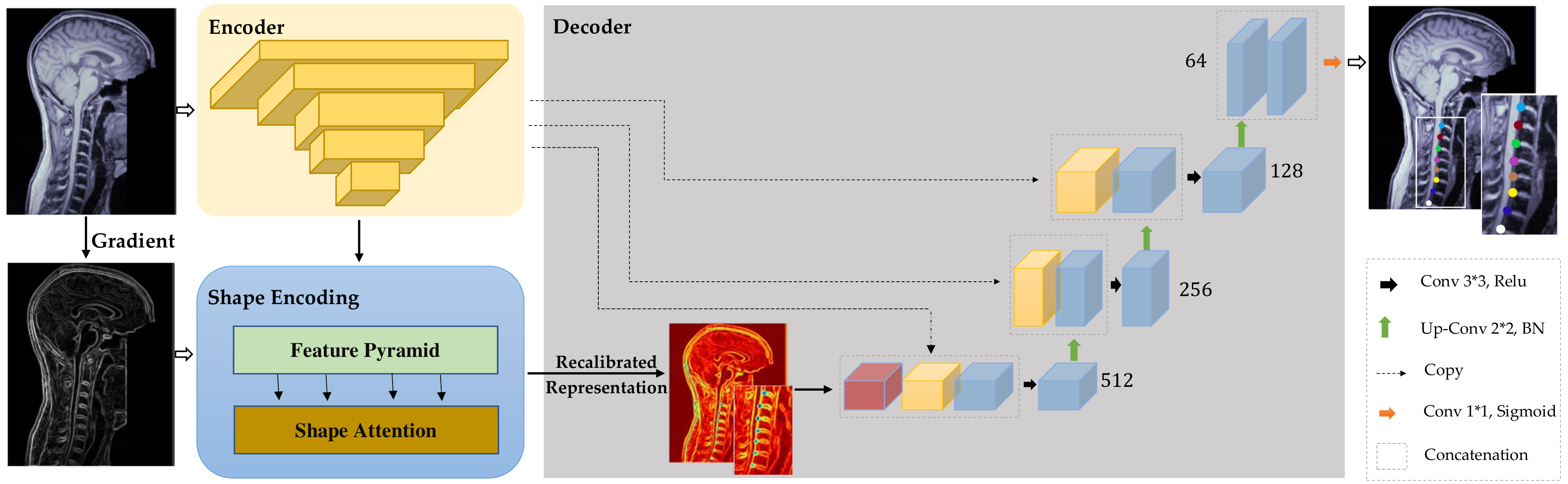}&
	\end{tabular}
	%\vspace{-4mm}
	\caption{Proposed method for intervertebral disc labeling with incorporating shape information.}
	%%\vspace{-2mm}
\end{figure}

\begin{equation}
w_{j}^{f}=\sigma\left(\mathbf{W}_{2} \delta\left(\mathbf{W}_{1} G A P_{j}^{f}\right)\right), 
w_{sp}=\sigma\left(\mathbf{W}_{4} \delta\left(\mathbf{W}_{3} G A P(sf)\right)\right)
\end{equation}
where $W_{k}, k \in\{1, 2,3,4\}$ are the learning parameters that apply to the global representation (GAP) of each pyramid level, and $\delta$ and $\sigma$ stand for the ReLU and Sigmoid activations. We form the re-calibrated description by scaling both channel and spatial dimensions: $ \tilde{P}_{j}^{f}=w_{sp} \cdot (w_{j}^{f} \cdot P_{j}^{f} )+ sf$. Once the re-calibration performed, we aggregate the multi-level features in a nonlinear fashion (aggregation parameter $w_{prm}$) to produce a shape-attenuating description: 
\begin{equation}
f'=\sigma\left(\sum_{j=1}^{L} w_{prm}^j \tilde{ P}_{j}^{f}\right)
\end{equation}

\begin{figure}[ht]
\label{fig:fig2}
%\vspace{-5mm}
	\centering
	\begin{tabular}{cc}
		\includegraphics[width=0.9\textwidth]{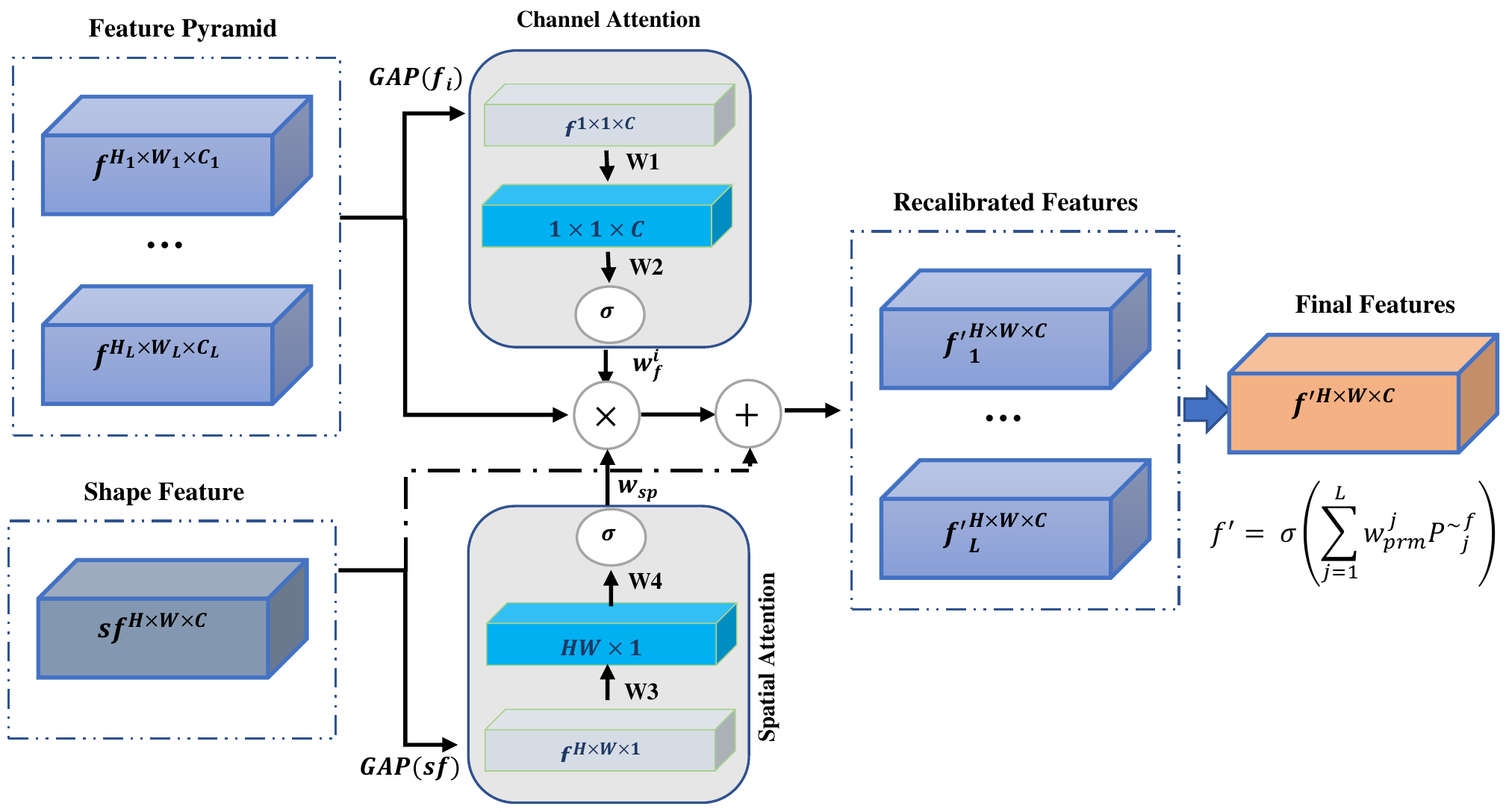}&
	\end{tabular}
	%\vspace{-4mm}
	\caption{Detailed structure of the proposed shape attention mechanism.}
	%%\vspace{-2mm}
\end{figure}

Subsequently, the same decoder as in the regular U-Net, but with $V=11$ output channels (we assume that the input image comprises, at most, 11 intervertebral discs according to \cite{spinedataset}), is utilized to estimate the location of each intervertebral disc accordingly. Similarly, our ground truth mask consists of V channels, where in each channel the location of an intervertebral disc is labelled with a Gaussian kernel of radius 10. We employ the mean squared (MSE) loss to train the network.

\subsection{Refinement Network}
\begin{figure}[t]
\begin{center}
  %\datasetylabel
  \includegraphics[width=1\linewidth]{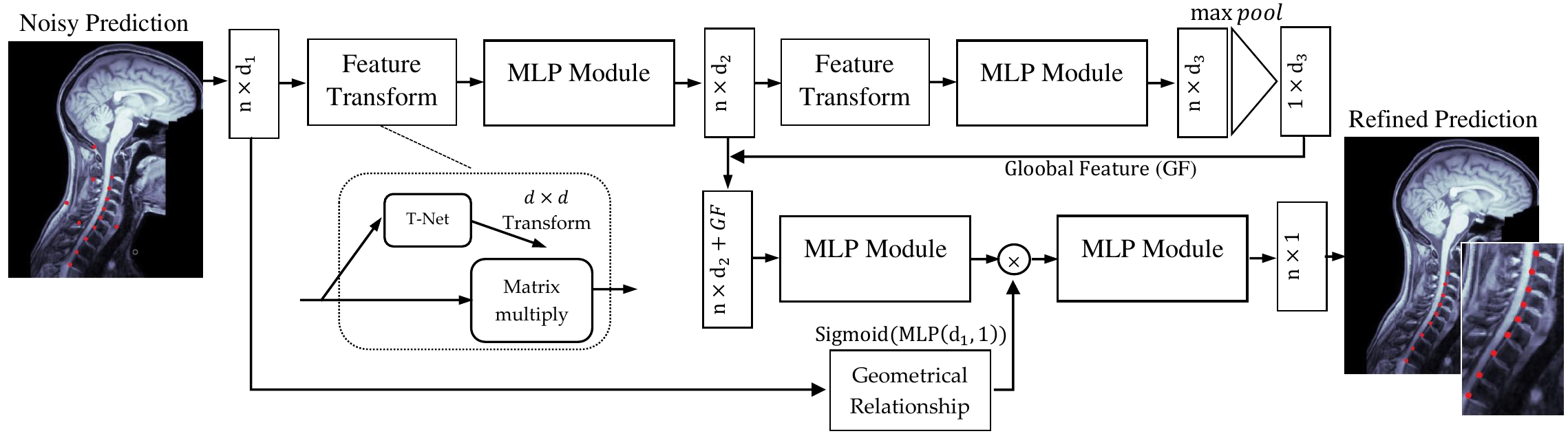}
\end{center}
   \caption{Proposed structure for the post-processing step. The noisy prediction without a semantic label passes through the model to eliminate the FP candidates.}
\label{fig:method.png}
\end{figure}

Detecting intervertebral disc locations often comprises FP and FN predictions. Several post-processing approaches were proposed in the literature to overcome this problem. Rouhier et al. 
\cite{rouhier2020spine}, deploys a condition-based strategy to eliminate the FP candidate generated by their countception method. In a recent article, Azad et al. \cite{azad2021stacked} argues that the condition-based strategy usually fails to recover the TP candidates among the detected regions and proposes a tree-based decision space. Their approach suggests creating a search tree, where each path shows one possible combination of ordered intervertebral disc locations. Then, they calculate an error function between the general skeleton and the predicted skeleton. This iterative algorithm performs an exhaustive search and is not efficient when the number of FP is high. Template matching \cite{ullmann2014automatic} is also another approach that seeks to reduce the FP rate by considering predefined patterns.\\
These methods all have their assumption of particular conditions or predefined patterns in common. In addition, some of these methods perform the selection in an iterative fashion, which may not be feasible when the number of FP is high. To mitigate these issues we propose a method to 'look only once' at the noisy prediction to recover the intervertebral disc locations. 
To this end, we assume that, for the input image I with N intervertebral disc location, the detection model predicts a set of $M$ intervertebral disc candidates, usually $M>=N$ and $M\in{R^2}$ (i.e. 2D position). Taking into further consideration in a general form, we assume that the prediction model is not able to provide any semantic labelling. Thus, the objective is to recover $N$ points out $M$ which best matches the ground truth intervertebral disc locations. Since the semantic information is not provided for the predicted points, we consider it as a set of $M$ intervertebral candidates. The set is made up of unstructured data and selecting $N$ intervertebral disc location out of $M$ candidates requires the following processing permutations:
\begin{equation}
\frac{(M)!}{(N)! (M-N)!} 
\end{equation}  
permutations. The processing time will dramatically increase if $M>>N$. To overcome this limitation, it is highly desirable that the post-processing algorithm processes the whole prediction at once without any iterations('look once'). Therefore, the deep model needs to be permutation invariant, i.e., any order of points should produce the same result. The proposed structure is depicted in Fig. \ref{fig:method.png}. 
The proposed method consists of two data streams, where in the first stream (top), a series of feature transformation layers, followed by the multilayer perceptron (MLP), is designed to encode the input coordinate into a high-level representational space. The objective of this representation is to create a discriminative embedding space to characterize each point by a hidden dependency underlying the input data. Intrinsically, the transformation layer in this stream assures the robustness of the representation to the noisy samples and provides a less sensitive transformation to an affine geometrical transformation (e.g. rotation). Inspired by the permutation invariance characteristics, the MLP layer deploys a shared kernel to produce a set of representations independent of their order. Eventually, in addition to the generated feature map, a symmetric function (global pooling) is utilized to capture the shared signature among all points. We concatenate the global information with the local representation of each point to describe each intervertebral disc candidate. This representation more or less contains the general structure of the data, however, it still requires pair-wise relational information. To include such information, we create a geometrical representation. To this end, using the fully connected layers, we learn the embedding parameters to model the long-range geometrical dependency. The main objective of this layer is to capture the geometrical relation between points and feed it to the scaler function. We include the sigmoid function on top of the generated representation to form an attention vector. This attention vector performs the re-calibration process and adaptively scales the generated feature map. The generated final representation is then fed to the single-layer perceptron model to perform the softmax operation and to classify each candidate.  

\section{Experimental Results}
In this section, we first describe the datasets and metrics used throughout our experimental evaluation. Then, we provide a deep insight into the experimental results. Our analysis was based on the publicly available Spine Generic Dataset \cite{spinedataset}. The dataset was acquired across 42 centers (with a total of 260 participants) worldwide, accommodating both T1 and T2 MRI contrasts for each subject. Images obtained from diverse institutes, considerably varying in image quality, ages and imaging devices, render a feasibly challenging benchmark for the task of intervertebral disc labelling.

\subsection{Metrics}
To ensure the validity of the comparison of results and to draw conclusions on the applicability of our approach, we consider different comparison metrics. In the first instance, we take into account the L2 norm by calculating the distance of the vector coordinate between each predicted intervertebral disc location and the ground truth while considering the superior-inferior axis to quantify the punctuality of our proposal. In order to gain insights into the versatility of our post-processing approach, the False Positive Rate (FPR) and False Negative Rate (FNR) were selected as the primary inclusion criteria. 
Similar to \cite{azad2021stacked}, the FPR calculates the number of predictions which are at least 5 mm away from the ground truth positions. Likewise, the FNR counts the number of predictions where the ground truth has at least 5mm distance from the predicted intervertebral position.

\subsection{Comparison of Results}

We train all of our models upstream using the Adam solver with the momentum in 100 epochs with the batch size 2. In our experiments, we use an initial learning rate of 0.0001 with the decay by a factor of 0.5 at every 20th epoch, respectively. We use the same setting
as explained in \cite{rouhier2020spine} to achieve a general consensus in comparing our method with the literature and we report our findings in Table 1.
Note: our baseline model uses the same structure as presented but without employing the proposed modules. 
The results show that our approach achieves a competitive result in T1 and T2 contrasts. Specifically, our proposed method shows superior performance in T2 contrast, where our approach prominently outperforms all other approaches in terms of FNR and distance to the target. Compared to the pose estimation approach \cite{azad2021stacked}, our method produces on T1 modality an average lower distance to the intervertebral locations, but there is only a small gap in distance variance. We also observe that, by removing the proposed modules the performance of the model slightly decreases, which highlights the importance of shape information in intervertebral disc labeling. Moreover, unlike the countception and template matching approaches, our method does not require a heavy preprocessing step for spinal cord region detection and outperforms these methods with both quantitative performance and inference time. In contrast to
our proposal, the inference time in the two aforementioned approaches grows exponentially when the FP rates increases. In Fig. 4(a) we provide sample results of the proposed model on T2 modalities. It can be observed that the method precisely provides a semantic label for each IVD location without any FP predictions.

\begin{table*}[t] %begin table top
	\caption{Intervertebral disc labeling results on the spine generic public dataset. Note that \textbf{DTT} indicates Distance to target}
	\resizebox{\textwidth}{!}{
		\begin{tabular}{c||c||c}
			\hline
			{\begin{tabular}{ccc}
		    	\multicolumn{3}{c}{\textbf{Method}}\\
					\hline
					\textbf{} & \textbf{} & \textbf{}\\
					\hline
			    Template Matching \cite{ullmann2014automatic}\\
				Countception \cite{rouhier2020spine}\\
				Pose Estimation \cite{azad2021stacked}\\
				Baseline\\
				\hline
				\multicolumn{3}{c}{\textbf{Proposed}}\\
				\hline
				\end{tabular}
			} &
			{\begin{tabular}{ccc}
				\multicolumn{3}{c}{\textbf{T1}}\\
					\hline
					\textbf{DTT (mm)} & \textbf{FNR (\%)} & \textbf{FPR (\%)}\\
					\hline
					1.97(±4.08) & 8.1 & 2.53 \\
					\textbf{1.03(±2.81)} & 4.24 & 0.9\\
					1.32(±1.33) & \textbf{0.32} & \textbf{0.0}\\
					1.45(±2.70) & 7.3 & 1.2\\
					\hline
					1.2(±1.90) & 0.7 & \textbf{0.0}\\
				\end{tabular}
			} &
			{\begin{tabular}{ccc}
					\multicolumn{3}{c}{\textbf{T2}}\\
					\hline
					\textbf{DTT (mm)} & \textbf{FNR (\%)} & \textbf{FPR (\%)}\\
					\hline
					2.05(±3.21) & 11.1 & 2.11\\
					1.78(±2.64) & 3.88 & 1.5\\
					1.31(±2.79) & 1.2 & 0.6\\
					1.80(±2.80) & 5.4 & 1.8\\
				    \hline
				    \textbf{1.28(±2.61)} & \textbf{0.9} & \textbf{0.0}\\
				\end{tabular}
			}
			\\
		\end{tabular}
	}
\end{table*}
%\vspace{-5mm}

\begin{figure}
\centering
\includegraphics[width=0.8\textwidth]{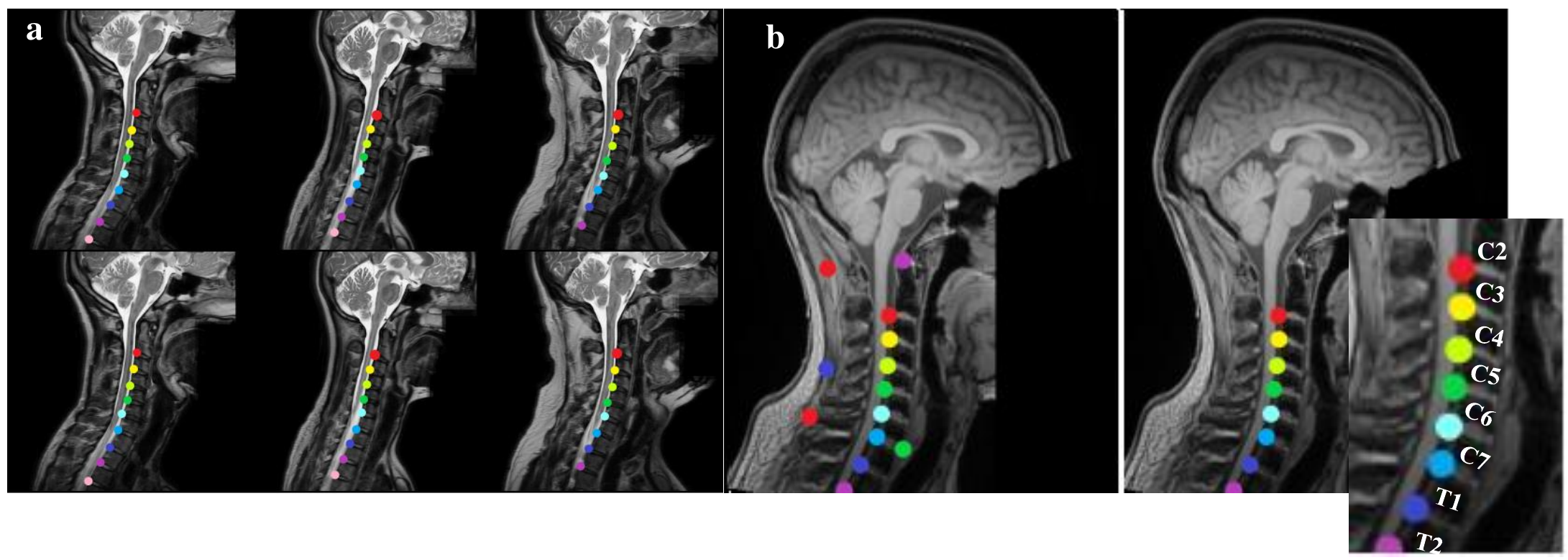}
\caption{(a): Intervertebral labeling results of three representative T2 images. upper row: ground truth, lower row: predictions. (b): Before (left) and after (right) applying look-once approach on the T1 generated noisy prediction.} \label{fig:4}
\end{figure}

\subsection{Evaluation on the Noisy Prediction}
To further analyze the robustness of the proposed method  in the presence of noisy predictions, we attain an evaluation on the proposed 'look once' post-processing method. To this end, we create a 2D Gaussian distribution around each intervertebral disc to generate new points. A sample of generated noisy image along with the model prediction is depicted in Fig. 4(b). As shown, the proposed method works well (including very fast timing) on retrieving IVD locations from the noisy prediction without relying on any predefined assumption. In addition, in our experiment (supplementary file), we observe that for the search-tree-based approach the post-processing time exponentially increased with the increase of FP rate. Similarly, the template matching method failed to recover the TP candidates in most of the cases. Whereas, our method recovered the TP samples with high precision without any iteration. Moreover, to disentangle the contribution of our proposal, we take a closer look at some additional sample detections of our method in the supplementary material section which proves its efficiency in terms of perceptual realism.

\section{Conclusion}
In this paper, we systematically formulate the intervertebral disc labelling problem by designing a novel method to incorporate shape information. The proposed method encourages the model to focus on learning contextual and geometrical features. Additionally, we propose a 'look once' post-processing approach. Powered by this, our model alleviates the false-positive detections along with a substantial refinement in model acceleration. The results presented in this paper demonstrate the potential of our methodology across all competing methods.

\bibliographystyle{splncs04}
\bibliography{Ref}

\begin{thebibliography}{10}
\providecommand{\url}[1]{\texttt{#1}}
\providecommand{\urlprefix}{URL }
\providecommand{\doi}[1]{https://doi.org/#1}

\bibitem{al2022transfer}
Al-kubaisi, A., Khamiss, N.N.: A transfer learning approach for lumbar spine
  disc state classification. Electronics  \textbf{11}(1), ~85 (2022)

\bibitem{asadi2020multi}
Asadi-Aghbolaghi, M., Azad, R., Fathy, M., Escalera, S.: Multi-level context
  gating of embedded collective knowledge for medical image segmentation. arXiv
  preprint arXiv:2003.05056  (2020)

\bibitem{ayed2011graph}
Ayed, I.B., Punithakumar, K., Garvin, G., Romano, W., Li, S.: Graph cuts with
  invariant object-interaction priors: application to intervertebral disc
  segmentation. In: Biennial International Conference on Information Processing
  in Medical Imaging. pp. 221--232. Springer (2011)

\bibitem{azad2019bi}
Azad, R., Asadi-Aghbolaghi, M., Fathy, M., Escalera, S.: Bi-directional
  convlstm u-net with densley connected convolutions. In: Proceedings of the
  IEEE/CVF International Conference on Computer Vision Workshops. pp.~0--0
  (2019)

\bibitem{azad2020attention}
Azad, R., Asadi-Aghbolaghi, M., Fathy, M., Escalera, S.: Attention deeplabv3+:
  Multi-level context attention mechanism for skin lesion segmentation. In:
  European Conference on Computer Vision. pp. 251--266. Springer (2020)

\bibitem{azad2021deep}
Azad, R., Bozorgpour, A., Asadi-Aghbolaghi, M., Merhof, D., Escalera, S.: Deep
  frequency re-calibration u-net for medical image segmentation. In:
  Proceedings of the IEEE/CVF International Conference on Computer Vision. pp.
  3274--3283 (2021)

\bibitem{azad2021texture}
Azad, R., Fayjie, A.R., Kauffmann, C., Ben~Ayed, I., Pedersoli, M., Dolz, J.:
  On the texture bias for few-shot cnn segmentation. In: Proceedings of the
  IEEE/CVF Winter Conference on Applications of Computer Vision. pp. 2674--2683
  (2021)

\bibitem{azad2022medical}
Azad, R., Khosravi, N., Dehghanmanshadi, M., Cohen-Adad, J., Merhof, D.:
  Medical image segmentation on mri images with missing modalities: A review.
  arXiv preprint arXiv:2203.06217  (2022)

\bibitem{azad2021stacked}
Azad, R., Rouhier, L., Cohen-Adad, J.: Stacked hourglass network with a
  multi-level attention mechanism: Where to look for intervertebral disc
  labeling. In: International Workshop on Machine Learning in Medical Imaging.
  pp. 406--415. Springer (2021)

\bibitem{badarneh2021semi}
Badarneh, A., Abu-Qasmeih, I., Otoom, M., Alzubaidi, M.A.: Semi-automated spine
  and intervertebral disk detection and segmentation from whole spine mr
  images. Informatics in Medicine Unlocked  \textbf{27},  100810 (2021)

\bibitem{bozorgpour2021multi}
Bozorgpour, A., Azad, R., Showkatian, E., Sulaiman, A.: Multi-scale regional
  attention deeplab3+: Multiple myeloma plasma cells segmentation in
  microscopic images. arXiv preprint arXiv:2105.06238  (2021)

\bibitem{chen2015localization}
Chen, C., Belavy, D., Yu, W., Chu, C., Armbrecht, G., Bansmann, M., Felsenberg,
  D., Zheng, G.: Localization and segmentation of 3d intervertebral discs in mr
  images by data driven estimation. IEEE transactions on medical imaging
  \textbf{34}(8),  1719--1729 (2015)

\bibitem{chen2014automatic}
Chen, C., Xie, W., Franke, J., Grutzner, P., Nolte, L.P., Zheng, G.: Automatic
  x-ray landmark detection and shape segmentation via data-driven joint
  estimation of image displacements. Medical image analysis  \textbf{18}(3),
  487--499 (2014)

\bibitem{chen2021study}
Chen, J.C., Lan, T.P., Lian, Z.Y., Chuang, C.H.: A study of intervertebral disc
  segmentation based on deep learning. In: 2021 IEEE 4th International
  Conference on Knowledge Innovation and Invention (ICKII). pp. 85--87. IEEE
  (2021)

\bibitem{cheng2021automatic}
Cheng, Y.K., Lin, C.L., Huang, Y.C., Chen, J.C., Lan, T.P., Lian, Z.Y., Chuang,
  C.H.: Automatic segmentation of specific intervertebral discs through a
  two-stage multiresunet model. Journal of clinical medicine  \textbf{10}(20),
  ~4760 (2021)

\bibitem{spinedataset}
Cohen-Adad, J., et~al: Open-access quantitative mri data of the spinal cord and
  reproducibility across participants, sites and manufacturers. sci. data. doi:
  10.1038/s41596-021-00588-0

\bibitem{dolz2018ivd}
Dolz, J., Desrosiers, C., Ayed, I.B.: Ivd-net: Intervertebral disc localization
  and segmentation in mri with a multi-modal unet. In: International workshop
  and challenge on computational methods and clinical applications for spine
  imaging. pp. 130--143. Springer (2018)

\bibitem{feyjie2020semi}
Feyjie, A.R., Azad, R., Pedersoli, M., Kauffman, C., Ayed, I.B., Dolz, J.:
  Semi-supervised few-shot learning for medical image segmentation. arXiv
  preprint arXiv:2003.08462  (2020)

\bibitem{glocker2012automatic}
Glocker, B., Feulner, J., Criminisi, A., Haynor, D.R., Konukoglu, E.: Automatic
  localization and identification of vertebrae in arbitrary field-of-view ct
  scans. In: International Conference on Medical Image Computing and
  Computer-Assisted Intervention. pp. 590--598. Springer (2012)

\bibitem{ji2016automated}
Ji, X., Zheng, G., Belavy, D., Ni, D.: Automated intervertebral disc
  segmentation using deep convolutional neural networks. In: International
  Workshop on Computational Methods and Clinical Applications for Spine
  Imaging. pp. 38--48. Springer (2016)

\bibitem{mbarki2020lumbar}
Mbarki, W., Bouchouicha, M., Frizzi, S., Tshibasu, F., Farhat, L.B., Sayadi,
  M.: Lumbar spine discs classification based on deep convolutional neural
  networks using axial view mri. Interdisciplinary Neurosurgery  \textbf{22},
  100837 (2020)

\bibitem{qi2017pointnet}
Qi, C.R., Su, H., Mo, K., Guibas, L.J.: Pointnet: Deep learning on point sets
  for 3d classification and segmentation. In: Proceedings of the IEEE
  conference on computer vision and pattern recognition. pp. 652--660 (2017)

\bibitem{redmon2016you}
Redmon, J., Divvala, S., Girshick, R., Farhadi, A.: You only look once:
  Unified, real-time object detection. In: Proceedings of the IEEE conference
  on computer vision and pattern recognition. pp. 779--788 (2016)

\bibitem{reza2022contextual}
Reza, A., Moein, H., Yuli, W., Dorit, M.: Contextual attention network:
  Transformer meets u-net. arXiv preprint arXiv:2203.01932  (2022)

\bibitem{rouhier2020spine}
Rouhier, L., Romero, F.P., Cohen, J.P., Cohen-Adad, J.: Spine intervertebral
  disc labeling using a fully convolutional redundant counting model. arXiv
  preprint arXiv:2003.04387  (2020)

\bibitem{ullmann2014automatic}
Ullmann, E., Pelletier~Paquette, J.F., Thong, W.E., Cohen-Adad, J.: Automatic
  labeling of vertebral levels using a robust template-based approach.
  International journal of biomedical imaging  \textbf{2014} (2014)

\bibitem{urban2003degeneration}
Urban, J.P., Roberts, S.: Degeneration of the intervertebral disc. Arthritis
  Res Ther  \textbf{5}(3),  1--11 (2003)

\bibitem{vania2021intervertebral}
Vania, M., Lee, D.: Intervertebral disc instance segmentation using a
  multistage optimization mask-rcnn (mom-rcnn). Journal of Computational Design
  and Engineering  \textbf{8}(4),  1023--1036 (2021)

\bibitem{wimmer2018fully}
Wimmer, M., Major, D., Novikov, A.A., B{\"u}hler, K.: Fully automatic
  cross-modality localization and labeling of vertebral bodies and
  intervertebral discs in 3d spinal images. International journal of computer
  assisted radiology and surgery  \textbf{13}(10),  1591--1603 (2018)

\end{thebibliography}

\newpage
\section{Appendix}

In this section, we provide some additional details regarding our approach, which allow a deeper understanding into our experiments.

\begin{figure}[ht]
\label{fig:fig1}
%\vspace{-5mm}
	\centering
	\begin{tabular}{cc}
		\includegraphics[width=1\textwidth]{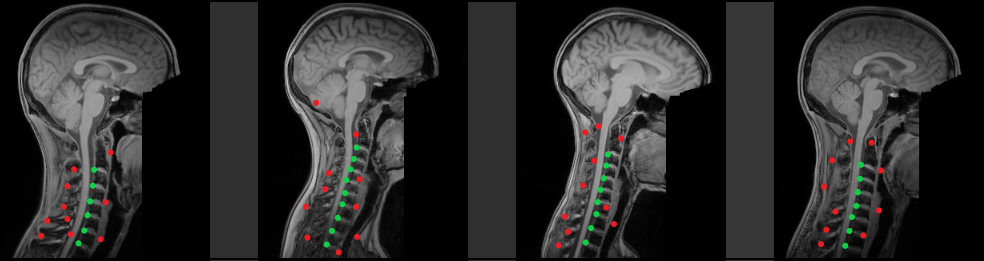}&
	\end{tabular}
	%\vspace{-4mm}
	\caption{Perceptual visualization of noisy samples generated for testing our post-processing approach (Note that 10 false positive (colored by red) samples were added as noise).}
	%%\vspace{-2mm}
\end{figure}

\begin{table}[h]
\caption{Performance comparison of the proposed post-processing approach vs the SOTA approach for eliminating FP detection. The experiment was done on 100 images, where for each image 20 random FP detection was added.}\label{table3}
\resizebox{1\textwidth}{!}{
\begin{tabular}{l||ccccc}
\hline Method & F1 & Accuracy &  specificity & sensitivity & AUC \\
\hline
Condition based (Rouhier et. al. \cite{rouhier2020spine}) & 0.850& 0.881 & 0.891 & 0.902 & 0.890\\

Search tree (Azad et. al. \cite{azad2021stacked}) & 0.902& 0.921 & 0.925 & 0.914 & 0.920\\

\hline
\hline 
Proposed method (without geometrical relationship module) & 0.914& 0.932 & 0.941 & 0.917 & 0.929\\ 

\textbf{Proposed method (Only look once)} & \textbf{0.942}& \textbf{0.958} & \textbf{0.967} & \textbf{0.942} & \textbf{0.955}\\ 
\end{tabular}
}
\end{table}

\begin{figure}[h]
\label{fig:fig1}
%\vspace{-5mm}
	\centering
	\begin{tabular}{cc}
		\includegraphics[width=1\textwidth]{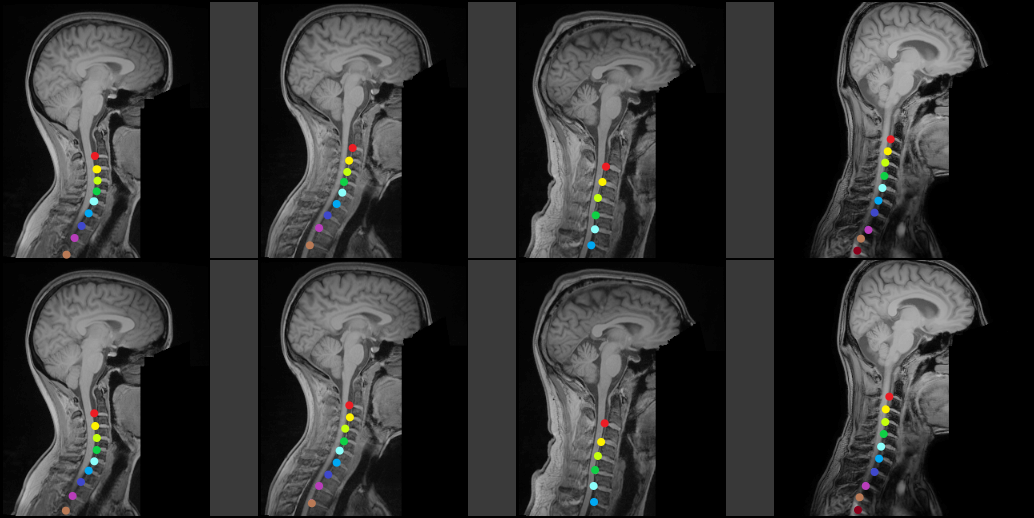}&
	\end{tabular}
	%\vspace{-4mm}
	\caption{More results of the proposed method for intervertebral disc labeling on T1w images. The first row shows the grand truth while the second row shows the predicted intervertebral disc along with the semantic labeling (color).}
	%%\vspace{-2mm}
\end{figure}

\begin{figure}[h]
\label{fig:fig1}
%\vspace{-5mm}
	\centering
	\begin{tabular}{cc}
		\includegraphics[width=1\textwidth]{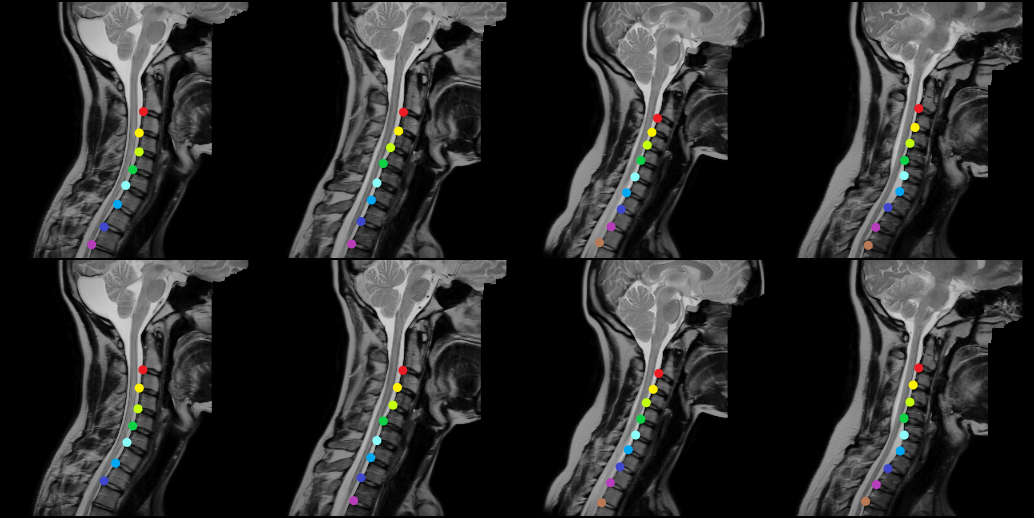}&
	\end{tabular}
	%\vspace{-4mm}
	\caption{More results of the proposed method for intervertebral disc labeling on T2w images. The first row shows the grand truth while the second row shows the predicted intervertebral disc along with the semantic labeling (color).}
	%%\vspace{-2mm}
\end{figure}

\begin{figure}[h]
\label{fig:fig1}
%\vspace{-5mm}
	\centering
	\begin{tabular}{cc}
		\includegraphics[width=1\textwidth]{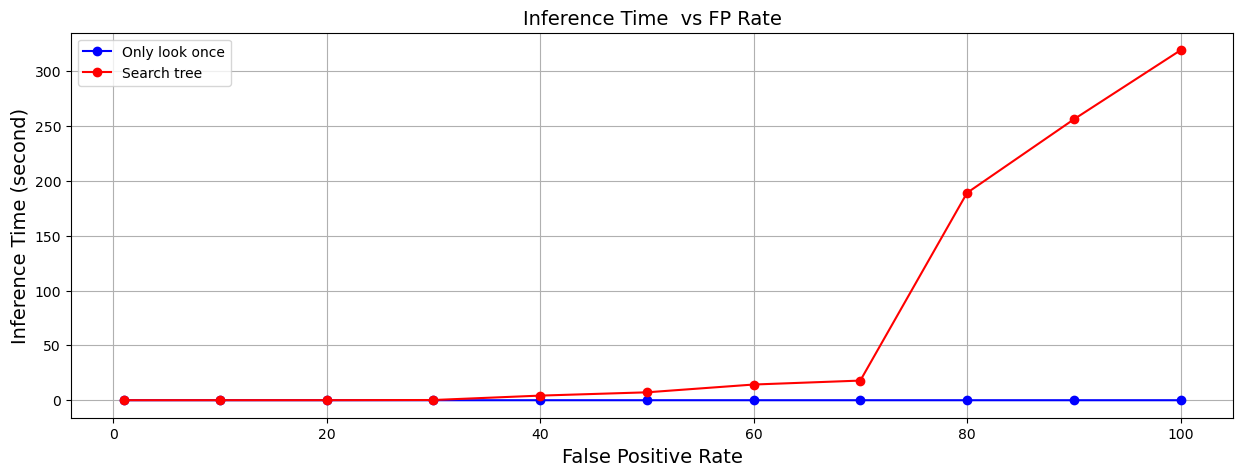}&
	\end{tabular}
	%\vspace{-4mm}
	\caption{Inference time of the proposed method vs the search-tree based approach. Our method only looks once at the prediction to eliminate the FP samples while the search based approach uses an iterative algorithm.}
	%%\vspace{-2mm}
\end{figure}

\end{document}